\documentclass[runningheads]{llncs}
\usepackage[T1]{fontenc}
\usepackage{amsmath}
\usepackage{graphicx}
\usepackage{booktabs}
\usepackage{multirow}
\usepackage{array,comment}
\usepackage{afterpage}
\usepackage{booktabs}
\usepackage{arydshln}
\usepackage{amssymb}
\usepackage{hyperref}
\usepackage[table,xcdraw]{xcolor}
\begin{document}
\title{Depth Attention for Robust RGB Tracking  }
%
%
\author{Yu Liu\inst{1}\orcidID{0009-0009-5898-0113} \and
Arif Mahmood\inst{2}\orcidID{0000-0001-5986-9876} \and
Muhammad Haris Khan\inst{3}\orcidID{0000-0001-9746-276X}\\ \textcolor{blue}{\textbf{Oral Acceptance at the Asian Conference on Computer Vision (ACCV) 2024, Hanoi, Vietnam.}}
}
\authorrunning{ Yu Liu et al.}
%
\institute{ Xinjiang University\\
\email{750184785ly@gmail.com}
\and
Information Technology University\\
\email{arif.mahmood@itu.edu.pk}
\and
Mohamed bin Zayed University of Artificial Intelligence
\email{muhammad.haris@mbzuai.ac.ae}
}
\maketitle              
\begin{abstract}
RGB video object tracking is a fundamental task in computer vision. Its effectiveness can be improved using depth information, particularly for handling motion-blurred target. However, depth information is often missing in commonly used tracking benchmarks. In this work, we propose a new framework that leverages monocular depth estimation to counter the challenges of tracking targets that are out of view or affected by motion blur in RGB video sequences. Specifically, our work introduces following contributions. To the best of our knowledge, we are the first to  propose a depth attention mechanism and to formulate a simple framework that allows seamlessly integration of depth information with state of the art tracking algorithms, without RGB-D cameras, elevating accuracy and robustness. We provide extensive experiments on six challenging tracking benchmarks. Our results demonstrate that our approach provides consistent gains over several strong baselines and achieves new SOTA performance. We believe that our method will open up new possibilities for more sophisticated VOT solutions in real-world scenarios. Our code and models are publicly released: \href{https://github.com/LiuYuML/Depth-Attention}{https://github.com/LiuYuML/Depth-Attention}.

\keywords{Visual Object Tracking \and Multi-Modal Tracking \and Monocular Depth Estimation \and Single object tracking.}
\end{abstract}
\section{Introduction}
\label{sec:intro}
RGB video object tracking\cite{bolme2010visual,henriques2014high,li2019siamrpn++,chen2021transformer} is a fundamental task in computer vision, aimed at accurately tracking a target object across video frames. Despite advancements in score map generation\cite{bolme2010visual,henriques2014high,bertinetto2016staple} and scale adaptation techniques \cite{danelljan2014accurate,bertinetto2016fully,danelljan2017eco,zhu2018distractor,li2019siamrpn++}, current SOTA algorithms\cite{paul2022robust,wang2019fast,lukezic2020d3s,gao2022aiatrack,fu2021stmtrack,Wei_2023_CVPR} face challenges in scenarios with invisible target, motion blur, and fast motion. 


\begin{figure}[!htp]
	\centering
	\includegraphics[width=1.0\linewidth]{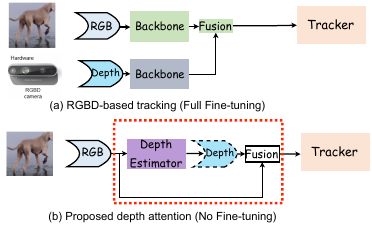}
	\caption{Comparison of our approach (depth attention) with RGB-D Tracking.}
	\label{teaser_pic}
\end{figure}
These challenges may be addressed  by employing depth information in the tracking algorithms. This integration makes the sudden object disappearances detectable through depth variation, while motion blur and fast motion have minimal impact on the depth pattern. However, incorporating RGB-D input for video tracking is usually costly and falls into the category of RGB-D video object tracking, as highlighted in various studies\cite{song2012tracking,yang2021rgbd,yan2021depthtrack,yang2023resource}. On the other hand, the present RGB datasets employed for RGB video object tracking lacks  depth information. To address this limitation and leverage the access to depth information in RGB Tracking, this paper introduces a simple tracking framework at the core of which is a new depth attention module specifically designed for RGB tracking algorithms. We would like to emphasize that the proposed module aims to enhance the performance of RGB tracking algorithms on the RGB datasets. Since the estimated monocular depth information is inherently less precise than the depth information from RGB-D sensors, we choose not to modify the architecture of existing RGB tracking methods as it is in the existing RGB-D algorithms \cite{kart2018make,kart2019object,liu2018context,VIPT}. Instead, we present a simple framework that can be seamlessly integrated into existing RGB tracking algorithms, resulting in improved performance on RGB datasets (see Figure~\ref{teaser_pic}). We summarize our key contributions as follows:

\begin{itemize}

\item 
To the best of our knowledge, we are the first to leverage depth information for improving RGB Tracking in a principled manner. 

\item Our approach is neither dependent on RGB-D datasets nor requires precise depth information from the RGB-D sensors. Our proposed depth attention efficiently leverages rapid monocular depth estimation and can be seamlessly incorporated into  existing RGB Tracking algorithms. 


\item 
The effectiveness of our proposed method is validated on six challenging benchmarks. The results consistently demonstrate the state-of-the-art (SOTA) performance across all RGB benchmarks.

\end{itemize}
\section{Related Work}
\noindent\textbf{RGB Visual Object Tracking:} RGB Visual Object Tracking (VOT) has been a vibrant area of research in computer vision, with a rich history of developments and a diverse array of approaches. The goal in RGB VOT is to accurately recover the trajectory of a target object within a sequence of frames, often under various challenges such as occlusions, illumination changes, and scale variations.

The evolution of RGB VOT algorithms has seen a transition from traditional methods like Mean Shift \cite{Comaniciu2003} and CamShift \cite{Bradski2000} to more sophisticated approaches leveraging machine learning and deep learning. Correlation filters, such as the Discriminative Correlation Filter (DCF) \cite{Henriques2012} and its Kernelized version (KCF) \cite{Henriques2014}, have significantly improved tracking performance by learning robust representations of the target object. Siamese networks, exemplified by SiamFC \cite{Bertinetto2016} and SiamRPN \cite{Li2018}, have introduced a novel way to track objects by learning to match target appearances across frames. Discriminative models, including CSRDCF \cite{Danelljan2014} and DiMP \cite{Bhat2019}, have further enhanced tracking by distinguishing the target from the background using convolutional neural networks (CNNs). Lately, the introduction of Vision Transformers (ViTs) \cite{Dosovitskiy2021} has revolutionized the field, with models like Mixformer \cite{Cui2022} and SeqTrack \cite{Chen2023} demonstrating the ability to handle long-range dependencies and complex patterns, leading to state-of-the-art tracking performance on several benchmarks.


\noindent\textbf{RGB-D Tracking:}. RGB-D tracking represents a significant advancement in the realm of visual object tracking (VOT), capitalizing on the depth data provided by RGB-D sensors to bolster tracking accuracy. The Princeton Tracking Benchmark\cite{song2012tracking} has played a pivotal role in the evolution of this domain, establishing a robust platform for assessing both RGB and RGB-D tracking algorithms. This benchmark comprises 100 diverse RGB-D video datasets, encompassing a broad spectrum of tracking challenges, and has enabled a systematic evaluation of various tracking methodologies\cite{song2013tracking}.

The integration of depth information has been proven to markedly enhance tracking capabilities, especially in situations of occlusions and model drift. The work of Ding et al.\cite{ding2015using}, have highlighted the benefits of incorporating depth images captured by devices like the Microsoft Kinect to refine conventional tracking techniques. The Depth Masked Discriminative Correlation Filter (DM-DCF)\cite{kart2018depth}, introduced by Kart et al., exemplifies an RGB-D tracking algorithm that leverages depth segmentation for occlusion detection and dynamically adjusts the spatial support for correlation filters, achieving high performance in the Princeton RGBD Tracking Benchmark.

A recent trend is to fuse depth and infra red (IR) imaging information to build a multimodal tracker\cite{VIPT}. However, none of these methods can be directly applied to the RGB datasets.

\noindent\textbf{Attention mechanisms:}. 
The introduction of attention mechanism in computer vision aims to mimic the capability of allowing models to dynamically weight features based on their importance, thereby enhancing performance across a variety of visual tasks.
Over time, attention mechanisms, categorized into channel, spatial, temporal, and branch attention \cite{Guo2022}, operate in various domains: channel for feature selection, spatial for region focus, temporal for time-series analysis, and branch for selecting among model components.

Channel attention, like in Squeeze-and-Excitation (SE) Networks \cite{Hu2018}, boosts feature importance through channel weight recalibration. Spatial attention, exemplified by Deformable Convolutional Networks (Deformable ConvNets) \cite{Dai2017}, selectively focuses on regions of interest for enhanced feature representation. Temporal attention, as seen in the Temporal Adaptive Module (TAM) \cite{Liu2020}, captures intricate temporal relationships for improved video recognition. Branch attention, demonstrated in Highway Networks \cite{Srivastava2015}, enables dynamic information flow across layers, addressing the vanishing gradient problem in deep networks.
In conclusion, attention mechanisms have become an indispensable tool in computer vision, offering a promising avenue for improving model performance and interpretability. As research continues, the integration of attention with other deep learning techniques and the development of more efficient and versatile attention models are expected to further advance the field. 

\noindent 
In this work, we introduce a novel approach to RGB Tracking by integrating depth information estimated from the RGB images. Our approach addresses some limitations of the traditional RGB-only tracking. We achieve this integration through a novel depth attention mechanism, offering a depth prior for the tracking task without requiring expensive RGB-D cameras or retraining the model.

\begin{figure}[!htp]
	\centering
	\includegraphics[width=1.0\linewidth]{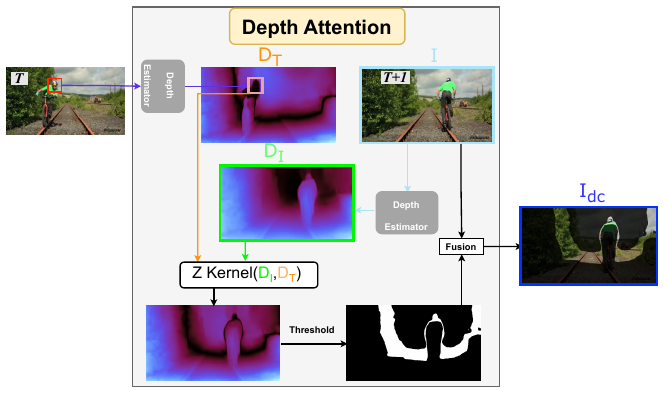}
	\caption{Proposed depth attention to improve RGB tracking.} 
	\label{intro_pic}
\end{figure}

\section{Depth Attention Based Tracking Framework}
We present a simple framework at the core of which is a new module that utilizes a rapid monocular depth estimation algorithm to create an initial depth map. Towards addressing potential errors stemming from the estimation process, we further refine this depth information using a novel Z kernel, distinguishing our proposed method from the current RGB-D Tracking pipeline\cite{yang2023resource}. The outcome is a probability map serving as a valuable prior for the tracking algorithms. Notably, this seamless integration into the existing RGB tracking algorithms demonstrates its adaptability and wide applicability. Figure \ref{intro_pic} sketches a schematic diagram of our proposed method.

\subsection{Monocular Depth Estimation}
Our monocular depth estimation approach leverages the Lite-Mono architecture \cite{Zhang_2023_CVPR}, a lightweight hybrid of CNNs and Transformers for self-supervised learning. This method is tailored for applications where stereo or LiDAR data is not available, enabling depth estimation from single images. The Lite-Mono model employs Consecutive Dilated Convolutions (CDC) to capture multi-scale local features and Local-Global Features Interaction (LGFI) to encode global context, which is crucial for accurate depth inference. The training objective combines photometric reprojection loss and edge-aware smoothness loss to ensure visually coherent and spatially consistent depth maps. The photometric reprojection loss is defined as \cite{Zhang_2023_CVPR}:
\begin{equation}
	\begin{aligned}
		L_r(\hat{I}_t, I_t) = \alpha/2  + (1 - \alpha) \cdot \|\hat{I}_t - I_t\| -\alpha SSIM(\hat{I}_t, I_t)/2
	\end{aligned}
	\label{eq001}
\end{equation} 
where $ \alpha $ is typically set to 0.85, and $ SSIM $ denotes the Structural Similarity Index. The edge-aware smoothness loss encourages spatial consistency in the predicted depth map $ d_t $ as follows \cite{Zhang_2023_CVPR}:
\begin{equation}
	\begin{aligned}
		L_{smooth} = \left| \frac{\partial d^*_t}{\partial x} \right| e^{-|\frac{\partial I_t}{ \partial x}|} + \left| \frac{\partial d^*_t}{\partial y} \right| e^{-|\frac{\partial I_t}{\partial y}|} ,
	\end{aligned}
	\label{eq002}
\end{equation} 
where $ d^*_t = \frac{d_t}{\hat{d}_t} $ is the mean-normalized inverse depth, and $ \frac{\partial}{\partial x} $ and $ \frac{\partial}{\partial y} $ represent the spatial gradients. The total loss function is a weighted sum of these two losses:
\begin{equation}
	\begin{aligned}
		L = \frac{1}{3} \sum_{s \in \{1, 1/2, 1/4\}} (L_r + \lambda L_{smooth}),
	\end{aligned}
	\label{eq002_2}
\end{equation} 
where $s$ is the output of depth decoder at different scales, and
$\lambda$ is a hyper-parameter. This approach allows for efficient and accurate depth estimation, making it suitable for real-time applications.
\subsection{Z Kernel and Signal Modulation}
Given a single RGB image, monocular depth estimation provides a single-channel score map, where each pixel value represents an estimated depth. However, this raw depth map is not directly suitable for existing RGB tracking algorithms, as these algorithms typically operate with input channels limited to 3. To address this, we propose our custom $Z_K$ kernel to create a probability map that highlights the region of interest within the bounding box:
\begin{equation}
   Z_K(D_I, D_T) = \left|\frac{(D_I - \text{ME}(D_T))}{\text{ME}(|D_T - \text{ME}(D_T)|)}\right|,
\end{equation}
where  $ D_I $ is the depth matrix of the entire current frame, while $ D_T $ denotes the depth matrix confined to the bounding box of the previous frame. The function $ \text{ME}(.) $ computes the median of the depth matrix. The $Z_K$ kernel measures the distance of $ D_I $ and $ D_T $. By setting a threshold $ Th $ on $Z_K$, a mask is obtained, focusing on the region of interest. Then we modulate the image with a linear formula:

\begin{figure}[!t]
	\centering
	\includegraphics[width=0.97\linewidth]{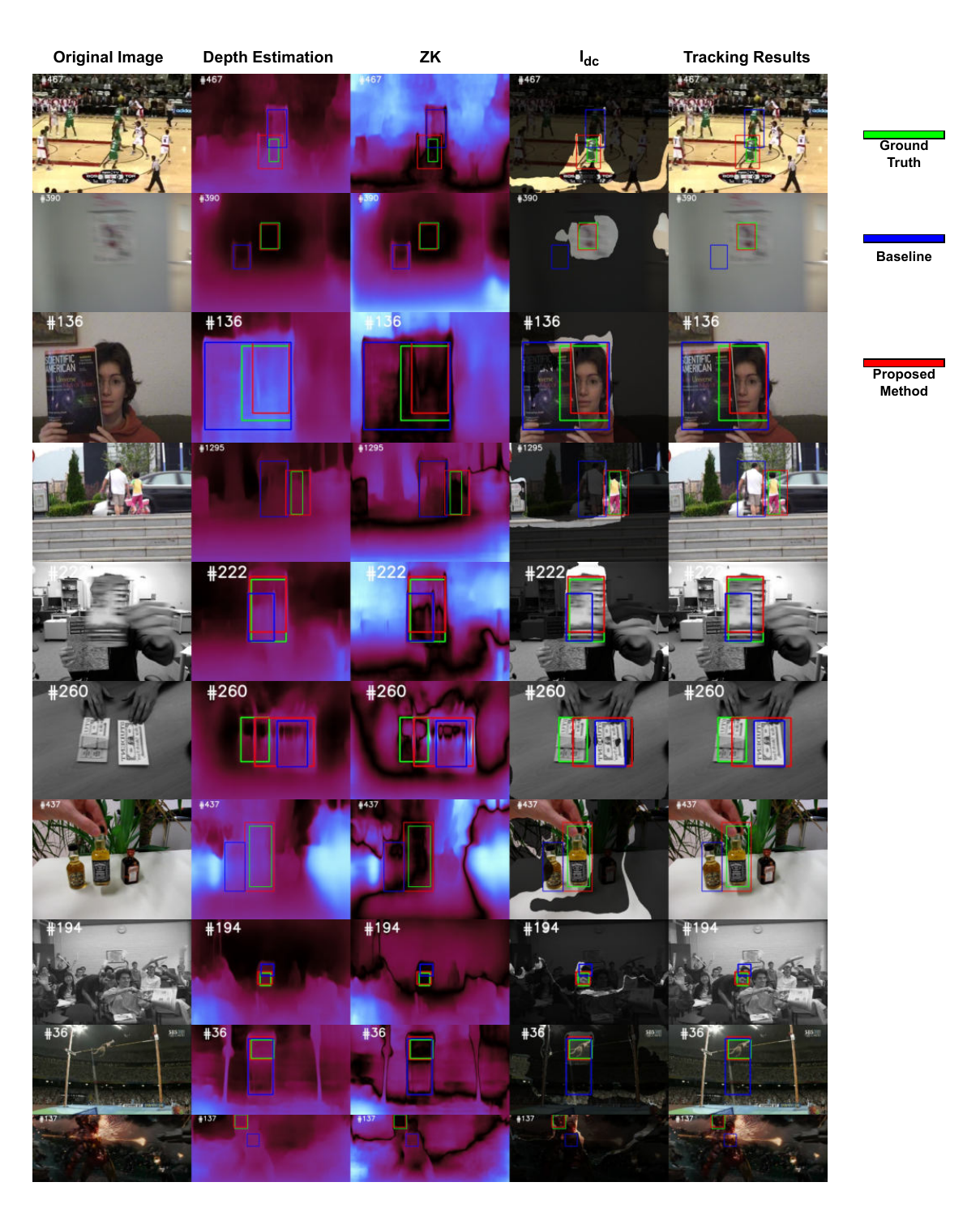}
	\caption{This figure shows the original images, their depth estimations, $Z_K$ values, I$_{dc}$, and tracking outcomes across four different sequences. Noticeably, when the baseline tracker starts to drift, our proposed method effectively prevents this by generating a mask with zero values in the background area. The proposed depth attention based masking  enhances the overall tracking performance.}
	\label{depth_estimation}
\end{figure}

\begin{equation}\label{eqx}
   I_{dc} = (1-k_1)  I + k_1  I \odot \text{Mask}(Z_K, Th),
\end{equation} where $\odot$ is element wise multiplication operator, and $k_1$ is the depth attention confidence. 
$I_{dc}$ represents the modulated image, the function $\text{Mask}(Z_K, th)$ transforms the depth map $Z_K$ into a binary mask based on the threshold $th$.  Balancing computational efficiency and precision, the D$_T$ undergoes periodic updates, specifically every 60 frames. Our method's periodic updates are controlled by statistical analysis, which indicates that across six key benchmarks, the target's movement is relatively small within certain frames, as further discussed in Section \ref{section:Experiments}. This insight anchors our strategy, ensuring that our approach maintains both adaptability and precision in the ever-changing landscape of visual tracking. The calculation for determining $k_1$ follows the formula: 
\begin{align}\label{eq11}
	k_1 = \frac{p(v_n;{\mu},{\sigma})}{\sum_{i=1}^{n}p(v_i;{\mu},{\sigma})},
\end{align} where  $v_i$ denotes the peak-to-sidelobe ratio (PSR) of the  confidence map, $p(\cdot)$ represents the Gaussian probability density function with parameters $\mu$ and $\sigma$. This formula is designed to calculate the proportion of the masked image. Instead of simply using the masked image on its own, we discovered that blending it with the original image significantly enhances the overall performance of our system. We would further verify this in Section \ref{section:Experiments}. $\mu$ and $\sigma$ are estimated from the collected PSR values $v_i, i=1, 2, 3, ..., n$. In the code, we specifically set $n=5$. For  better understanding, we draw the results in Figure \ref{depth_estimation}. This entire procedure can be conceptualized as a disentangling process for the target object from the background. Moreover, note that this method is compatible with all existing tracking methods.

\section{Experiments}
\label{section:Experiments}
\textbf{Datasets:}
Our approach has been rigorously validated on several benchmark datasets, which are widely recognized for their role in assessing the performance of visual tracking algorithms. 
\begin{itemize}
   \item \textbf{OTB100} \cite{7001050}: This dataset consists of 100 video sequences with a diverse range of tracking challenges, such as illumination changes, occlusions, and camera motion. It is one of the most popular benchmarks for evaluating tracking algorithms due to its comprehensiveness and difficulty.

   \item \textbf{NfS} \cite{kiani2017need}: The Need for Speed (NfS) dataset is designed to test tracking algorithms under high frame rates, capturing videos with high-speed motion and rapid scene changes. It pushes the boundaries of tracking algorithms by introducing extreme tracking scenarios.

   \item \textbf{AVisT} \cite{noman2022avist}: The AVisT dataset focuses on tracking under adverse visibility conditions, such as low-light and high clutter. It presents a significant challenge for trackers, as it requires robust tracking in scenarios where visual information is limited.

   \item \textbf{UAV123} \cite{mueller2016benchmark}: This dataset is specifically tailored for tracking small objects in aerial videos, often at high altitudes and with significant perspective changes. It introduces unique challenges due to the scale and motion of the objects in the context of aerial imagery.

   \item \textbf{LaSOT} \cite{fan2019lasot}: The LaSOT dataset is a large-scale single object tracking benchmark that includes a diverse set of 1,550 videos. It is designed to provide a comprehensive evaluation of tracking algorithms across various challenging conditions, such as occlusions, scale variations, and fast motion.

   \item \textbf{GOT-10k} \cite{huang2019got}: The GOT-10k dataset is a large-scale benchmark for generic object tracking in the wild, containing over 10,000 video clips with a wide variety of object classes and tracking scenarios. It is known for its high diversity and the complexity of the tracking tasks it presents.
    
   \item \textbf{NT-VOT211} \cite{ntvot211}: This dataset represents a novel benchmark, specifically designed for evaluating visual object tracking algorithms under challenging low-light conditions. It comprises 211 videos and 211k annotated frames. It poses the unique challenges inherent to the nighttime tracking scenarios. 

\end{itemize}


\noindent\textbf{Implementation Details:} All results in this paper were obtained within a consistent computational environment, utilizing Python 3.8.10, PyTorch 1.11.0 with CUDA 11.3, NumPy 1.22.3, and OpenCV 4.8.0. The Depth Estimation model was initially trained following the procedures outlined in \cite{Zhang_2023_CVPR}. The training process was conducted on a single Tesla V100 GPU, while subsequent evaluations were performed using a Nvidia A5000 GPU. Here, the sole hyperparameter $Th$ in equation \ref{eqx} was maintained at a constant value of $Th = 1.5$.

\noindent\textbf{Evaluation metrics:} We use a range of metrics, including AUC, success scores, and AO, tailored to each benchmark as specified in their respective papers\cite{7001050,fan2019lasot,huang2019got,kiani2017need}. NT-VOT211 dataset\cite{ntvot211} adheres to the evaluation protocol of LaSOT\cite{fan2019lasot}, ensuring a consistent and fair assessment of tracking performance.

\noindent\textbf{Results:} We integrate the proposed method into ODTrack \cite{ODTrack}, the SOTA on LaSOT \cite{fan2019lasot}, TrackingNet \cite{muller2018trackingnet}, TNL2k \cite{TNL2k}, and ITB \cite{itb}. We have listed all the results in Table \ref{SOTA1}. As shown, we have achieved new SOTA on all six benchmarks. The better performance highlighted in \textbf{\textcolor{red}{red} bold font}, and `+DA' indicates the integration of proposed Depth Attention (DA) module.

\begin{table}[] \vspace{-1em}
\caption{Comparison of state-of-the-art methods with and without the proposed Depth Attention (DA) module on six benchmarks: GOT10K, LaSOT, TrackingNet, TNL2k, UAV123, and NfS. The best results are highlighted in \textbf{\textcolor{red}{red}} and the second best in \textbf{\textcolor{blue}{blue}} colors.}
\label{SOTA1}
\begin{scalebox}{0.75}{
\begin{tabular}{l|llllll}
\toprule
                                  & GOT10k\cite{huang2019got}                      & LaSOT\cite{fan2019lasot} & TrackingNet\cite{muller2018trackingnet} & TNL2k\cite{kim2022towards} & UAV123\cite{mueller2016benchmark}                               & Nfs\cite{kiani2017need}   \\ \cline{2-7} 
\multirow{-2}{*}{Method}          & (AO)                        & (AUC) & (Success)   & (AUC) & (AUC)                                & (AUC) \\ \hline
TransT\cite{chen2021transformer} & 64.9                        & 67.1  & 81.4        & 50.7  & -                                    & -     \\
OSTrack\cite{ye2022joint}        & 73.7                        & 71.1  & 83.9        & 55.9  & -                                    & -     \\
SwinTrack\cite{swintrack}        & 72.4                        & 71.3  & 84.0        & -     & -                                    & -     \\
Mixformer-22k\cite{cui2022mixformer} & 70.7                        & 69.2  & 83.1        & -     & 69.5                                 & -     \\
SeqTrack-$B_{384}$\cite{chen2023seqtrack} & 74.5                        & 71.5  & 83.9        & -     & 68.6                                 & 66.7  \\
VideoTrack\cite{videotrack}      & 72.9                        & 70.2  & 83.8        & -     & -                                    & -     \\ \hdashline
RTS\cite{paul2022robust}         & 73.5                        & 69.7  & 81.6        & 54.7  & 67.6                                 & 65.4  \\
\textbf{RTS+DA (31 fps)}         & 74.2                        & 70.3  & 82.1        & 55.6  & 68.1                                 & 66.0  \\ \hdashline
DropTrack\cite{wu2023dropmae}    & 75.9                        & 71.8  & 84.1        & 57.7  & 69.5                                 & 66.7  \\
\textbf{DropTrack+DA (22 fps)}   & {\color[HTML]{000000} 77.0} & 72.4  & 84.6        & 58.2  & 70.4                                 & 67.2  \\ \hdashline
ARTrack\cite{wei2023autoregressive} & 76.6                        & 70.4  & 84.0        & 57.5  & 67.7 & 66.8  \\
\textbf{ARTrack+DA (21 fps)}     & {\color[HTML]{3166FF} \textbf{77.5}} & 70.7 & 84.2 & 57.7 & 68.1 & 67.4 \\ \hdashline
ODTrack\cite{ODTrack}           & 77.0                        & {\color[HTML]{3166FF} \textbf{73.4}}  & {\color[HTML]{3166FF} \textbf{85.7}} & {\color[HTML]{3166FF} \textbf{61.7}} & {\color[HTML]{3166FF} \textbf{69.5}} & {\color[HTML]{3166FF} \textbf{67.6}} \\
\textbf{ODTrack+DA (8fps)}       & {\color[HTML]{FE0000} \textbf{78.8}} & {\color[HTML]{FE0000} \textbf{74.3}} & {\color[HTML]{FE0000} \textbf{86.1}} & {\color[HTML]{FE0000} \textbf{61.9}} & {\color[HTML]{FE0000}\textbf{70.3}} & {\color[HTML]{FE0000} \textbf{68.5}} \\ \bottomrule
\end{tabular}}
\end{scalebox}
\end{table}

Next, we integrate the proposed Depth Attention (DA) module into 2023 SOTA and evaluate on different benchmarks. We have listed all the results in Table \ref{SOTA}. We observe a consistent performance improvement highlighted in \textbf{\textcolor{red}{red} bold font}. 

\begin{table}[ht]
	\centering
	\caption{More comparisons with SOTA methods. Addition of Depth Attention (DA) module has consistently improved performance. The evaluation measures Average Overlap (AO), Success (S), Precision (P), Normalized Precision (NP) are used; OP50 refers to the Overlap Precision at 50\% threshold and
OP75 refers to the Overlap Precision at 75\% threshold, SR50 refers to the Success Rate at a 50\% IoU threshold,
SR75 refers to the Success Rate at a 75\% IoU threshold. The 'rank' column denotes the ranking of each method on their corresponding datasets, with the specific metric used for ranking provided in parentheses.}\label{SOTA}
 \begin{scalebox}{0.65}{
	\begin{tabular}{@{}lccccccc@{}}
		\toprule
		Method &Dataset&rank&\multicolumn{5}{c}{Metric}\\
  \midrule
		\multirow{2}{*}{ARTrack-L \cite{wei2023autoregressive}(24 fps)} &\multirow{3}{*}{GOT-10k \cite{huang2019got}} & \multirow{2}{*}{3(AO)}&AO&  &SR50& &SR75\\
		& & & 76.6 & & 85.6& &75.4\\
		\multirow{1}{*}{\textbf{ARTrack-L+DA}(21 fps)} & & \multirow{1}{*}{\textcolor{red}{\textbf{2}}(AO)}& \textcolor{red}{\textbf{77.6}}&  &\textcolor{red}{\textbf{87.2}}&  &\textcolor{red}{\textbf{76.6}}\\
    \midrule
		\multirow{2}{*}{STMTrack \cite{fu2021stmtrack}(43 fps)} &\multirow{3}{*}{OTB100\cite{7001050}} & \multirow{2}{*}{2(S)} & S & && & P\\
		& & & 71.43 &   && &92.64\\
		\multirow{1}{*}{\textbf{STMTrack+DA}(39 fps)} &  & \multirow{1}{*}{\textcolor{red}{\textbf{1}}(S)}& \textcolor{red}{\textbf{71.88}}& &&&\textcolor{red}{\textbf{93.01}}\\
  \midrule
		\multirow{2}{*}{Neighbor-OSTrack \cite{Chen_2023_CVPR}(53 fps)} &\multirow{3}{*}{UAV123 \cite{mueller2016benchmark}} & \multirow{2}{*}{2(AUC)} &AUC& P& NP&OP50&OP75\\
		& & & 72.56 &93.37& 88.51&87.75&68.15\\
		\multirow{1}{*}{\textbf{Neighbor-OSTrack+DA}(49 fps)} &  & \multirow{1}{*}{\textcolor{red}{\textbf{1}}(AUC)}& \textcolor{red}{\textbf{72.99}}&\textcolor{red}{\textbf{94.12}} &\textcolor{red}{\textbf{89.00}}&\textcolor{red}{\textbf{89.01}} &\textcolor{red}{\textbf{68.78}}\\
  \midrule
		\multirow{2}{*}{Mixformer(ConvMAE)-L \cite{cui2022mixformer}(27 fps)} &\multirow{3}{*}{LaSOT \cite{fan2019lasot}} & \multirow{2}{*}{5(AUC)} &AUC&P& NP&OP50&OP75\\
		& & & 72.31 &78.93& 81.66&83.63&72.17\\
		\multirow{1}{*}{\textbf{Mixformer(ConvMAE)-L+DA}(25 fps)} & & \multirow{1}{*}{\textcolor{red}{\textbf{3}}(AUC)}& \textcolor{red}{\textbf{73.01}}&\textcolor{red}{\textbf{80.83}} &\textcolor{red}{\textbf{83.57}}&\textcolor{red}{\textbf{84.39}} &\textcolor{red}{\textbf{73.87}}\\
  \midrule
		\multirow{2}{*}{AiATrack \cite{gao2022aiatrack}(35 fps)} &\multirow{3}{*}{NfS \cite{kiani2017need}} & \multirow{2}{*}{2(AUC)} &AUC& P& NP&OP50&OP75\\
		& & & 68.40 &84.70& 87.29&87.37&\textcolor{red}{\textbf{55.08}}\\
		\multirow{1}{*}{\textbf{AiATrack+DA}(33 fps)} & & \multirow{1}{*}{\textcolor{red}{\textbf{1}}(AUC)}& \textcolor{red}{\textbf{69.27}}&\textcolor{red}{\textbf{85.38}} &\textcolor{red}{\textbf{88.98}}&\textcolor{red}{\textbf{88.84}} &55.05\\
  \midrule
		\multirow{2}{*}{KeepTrack \cite{mayer2021learning}(19 fps)} &\multirow{3}{*}{NT-VOT211 \cite{ntvot211}} & \multirow{2}{*}{2(P)} &AUC& P& NP&OP50&OP75\\
		& & & 39.59 &55.50& 85.06&50.52&\textcolor{red}{\textbf{12.83}}\\
		\multirow{1}{*}{\textbf{KeepTrack+DA}(15 fps)} & & \multirow{1}{*}{\textcolor{red}{\textbf{1}}(P)}& \textcolor{red}{\textbf{39.99}}&\textcolor{red}{\textbf{55.72}} &\textcolor{red}{\textbf{85.88}}&\textcolor{red}{\textbf{50.90}} &12.77\\
		\bottomrule
	\end{tabular}}
 \end{scalebox}
\end{table}

\noindent\textbf{Comparison with SOTA RGBD tracking methods:}
 First, we evaluate VIPT\cite{VIPT} (RGBD tracker) and ODTrack + DA (ours) on GOT10k. ODTrack\cite{ODTrack} is a recent SOTA on many datasets. To incorporate depth information into VIPT, a necessity for RGBD tracking algorithms, we utilize depth estimates from a monocular architecture to estimate the depth information of GOT-10k video. Second, we compare VIPT and VIPT + DA (ours) on DepthTrack dataset\cite{DepthTrack}, a RGBD dataset.  
 \begin{table}[!htp]
 \vspace{+2em}
\centering
\caption{Comparison with RGBD trackers.}
\label{tab:rgbd_comparison}
\scalebox{1.0}{
\begin{tabular}{|l|ll|l|ll|}
\hline
Strategy 1 &
  \multicolumn{2}{l|}{Method} &
  Strategy 2 &
  \multicolumn{2}{l|}{Method} \\ \hline
 &
  \multicolumn{1}{l|}{VIPT} &
  72.1 &
   &
  \multicolumn{1}{l|}{VIPT} &
  57.8 \\ \cline{2-3} \cline{5-6} 
\multirow{-2}{*}{\begin{tabular}[c]{@{}l@{}}GOT10k\\ (AO)\end{tabular}} &
  \multicolumn{1}{l|}{ODTrack+DA} &
  {\color[HTML]{FE0000} \textbf{78.8}} &
  \multirow{-2}{*}{\begin{tabular}[c]{@{}l@{}}DepthTrack\\ (F scores)\end{tabular}} &
  \multicolumn{1}{l|}{VIPT+DA} &
  {\color[HTML]{FE0000} \textbf{58.6}} \\ \hline
\end{tabular}}
\end{table}

\noindent\textbf{Ablation study on $k_1$}: As stated in Equation \ref{eqx}, the confidence of the depth attention mechanism significantly influences the extent to which the original image is modulated. We conducted experiments by varying the parameter $k_1$ across different constants in the RTS method \cite{paul2022robust}. See Tables \ref{otb_single} and Table \ref{rtsatt}.
\begin{table}
    \caption{Ablation study on $k_1$, We evaluated the RTS under different parameters on OTB100}
    \label{otb_single}
    \centering
    \scalebox{0.7}{ 
    \begin{tabular}{l|ccccccccccccc}
        \toprule
        Hyper-parameter & $k_1$ & 1.0 & 0.9 & 0.8 & 0.7 & 0.6 & 0.5 & 0.4 & 0.3 & 0.2 & 0.1 & 0.0 & 0.02 \\
        (Th=1.5) & & & & & & & & & & & & & \\
        \midrule
        AUC & & 34.60 & 56.47 & 59.13 & 61.55 & 63.14 & 64.87 & 64.40 & 64.60 & 65.45 & 66.09 & \textcolor{black}{66.17} &\textcolor{red}{\textbf{66.35}} \\
        Precision & & 43.63 & 75.15 & 79.14 & 82.85 & 84.66 & 87.02 & 85.92 & 86.04 & 87.10 & 88.14 & \textcolor{black}{87.94} & \textcolor{red}{\textbf{88.45}} \\
        Norm Precision & & 38.38& 65.41 & 69.11 & 72.82 & 74.29 & 76.52 & 75.93 & 75.77 & 76.75 & 77.74 & \textcolor{black}{77.66} & \textcolor{red}{\textbf{78.36}} \\
        \bottomrule
    \end{tabular}}\vspace{-2em}
\end{table}

\begin{table}[!htpb]
    \vspace{+2em}
    \caption{Optimal fixed $k_1$ and Th}
    \label{rtsatt}
    \centering
    \scalebox{0.7}{ 
    \begin{tabular}{l|cc|cc|cc|c|cc}
        \toprule
        Hyper-parameter&\multicolumn{2}{c}{OTB}&\multicolumn{2}{c}{LaSOT}& \multicolumn{2}{c}{AVisT} &  &\multicolumn{2}{c}{GOT-10k}\\
        \cline{1-10}
        (Th=1.5) &\multirow{2}{*}{RTS} &\multirow{2}{*}{RTS+DA} &\multirow{2}{*}{RTS} &\multirow{2}{*}{RTS+DA}&\multirow{2}{*}{RTS} &\multirow{2}{*}{RTS+DA}&  &\multirow{2}{*}{RTS} &\multirow{2}{*}{RTS+DA} \\
        ($k_1$=0.02) & & & & & & & & &  \\
        \cline{0-9}
        AUC & 66.17& \textcolor{red}{\textbf{66.35}} & 69.46 & \textcolor{red}{\textbf{70.04}} & 49.98 & \textcolor{red}{\textbf{50.81}} & AO &73.5& \textcolor{red}{74.1} \\
        Precision &87.94 & \textcolor{red}{\textbf{88.45}} & 73.21 & \textcolor{red}{\textbf{73.89}} & 47.78 & \textcolor{red}{\textbf{48.74}} & $SR_{0.5}$ &83.1&\textcolor{red}{\textbf{83.7}}\\
        Norm Precision & 77.6 & \textcolor{red}{\textbf{78.36}} & 76.05 & \textcolor{red}{\textbf{76.83}} & 65.46 & \textcolor{red}{\textbf{66.12}} &$SR_{0.75}$&68.3&\textcolor{red}{\textbf{68.7}}\\
        \bottomrule
    \end{tabular}}
\end{table}

\begin{figure}[]
	\centering
	\includegraphics[width=1.0\linewidth]{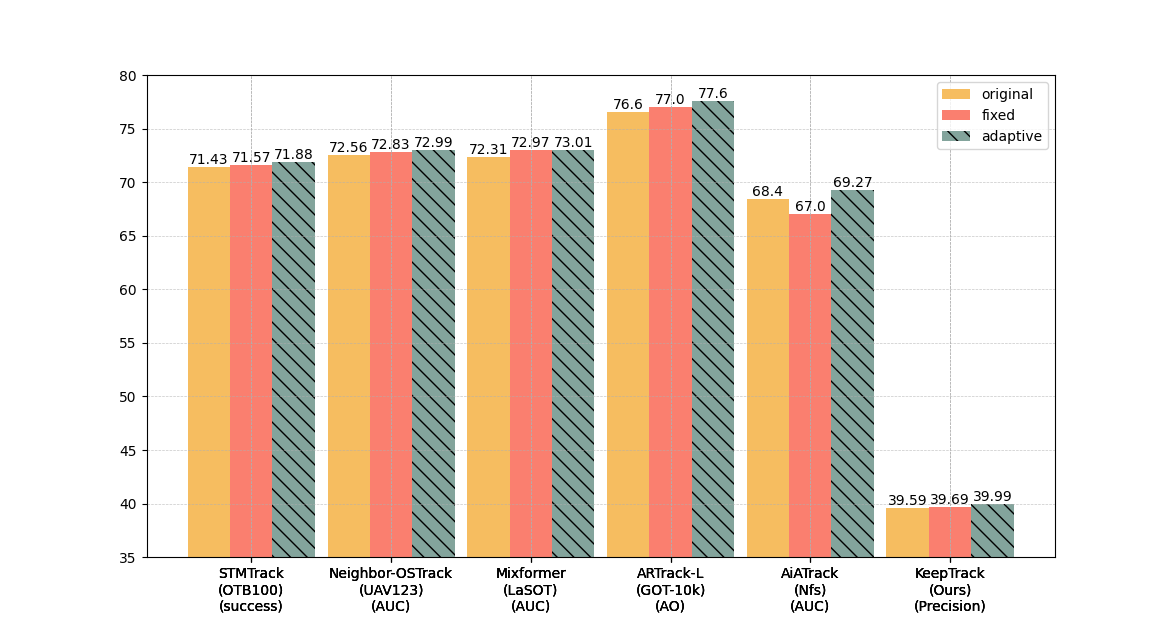}
	\caption{We compare the original algorithm, the depth attention module with a fixed hyperparameter (Th=1.5, $k_1=0.02$), and the depth attention module with an adaptive hyperparameter on $k_1$. We elaborate on the base algorithm, dataset, and the metrics used along the axes.}
	\label{barchar}
\end{figure}

Subsequently, we applied the optimal constant $k_1=0.02$ to the module and compared it with the proposed module, where $k_1$ is estimated using Equation \ref{eq11}. The results are illustrated in Figure \ref{barchar}. It is evident that the adaptive estimation process for $k_1$ plays a vital role in enhancing performance.

\begin{figure}[]
	\centering
	\includegraphics[width=1.0\linewidth]{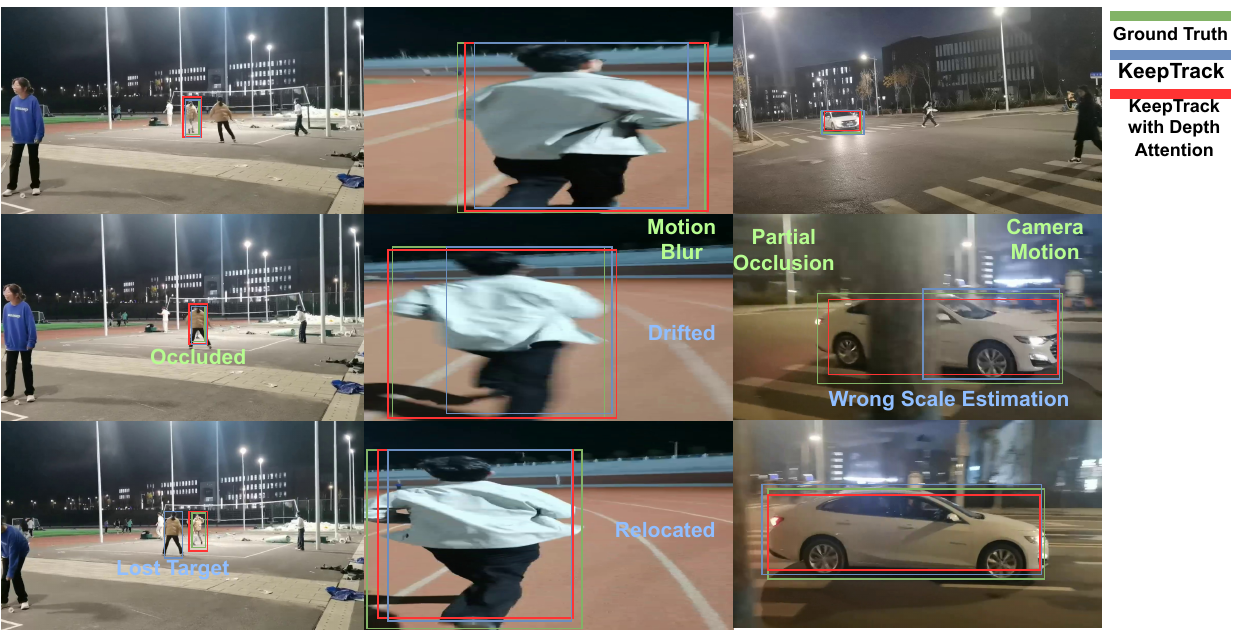}
	\caption{Visual comparison between KeepTrack and KeepTrack+DA (ours). It can be seen that, our method is robust under the challenging scenarios of occlusion and motion blur.}
	\label{add1}
\end{figure}

\noindent\textbf{Qualitative Results:} Figure \ref{add1}, illustrates the tracking process of KeepTrack\cite{mayer2021learning} and KeepTrack with our depth attention. Our proposed method is adept at handling challenging scenarios e.g., occlusion, motion blur, and camera motion.


\begin{figure}[!htp]
	\centering	\includegraphics[width=0.95\linewidth]{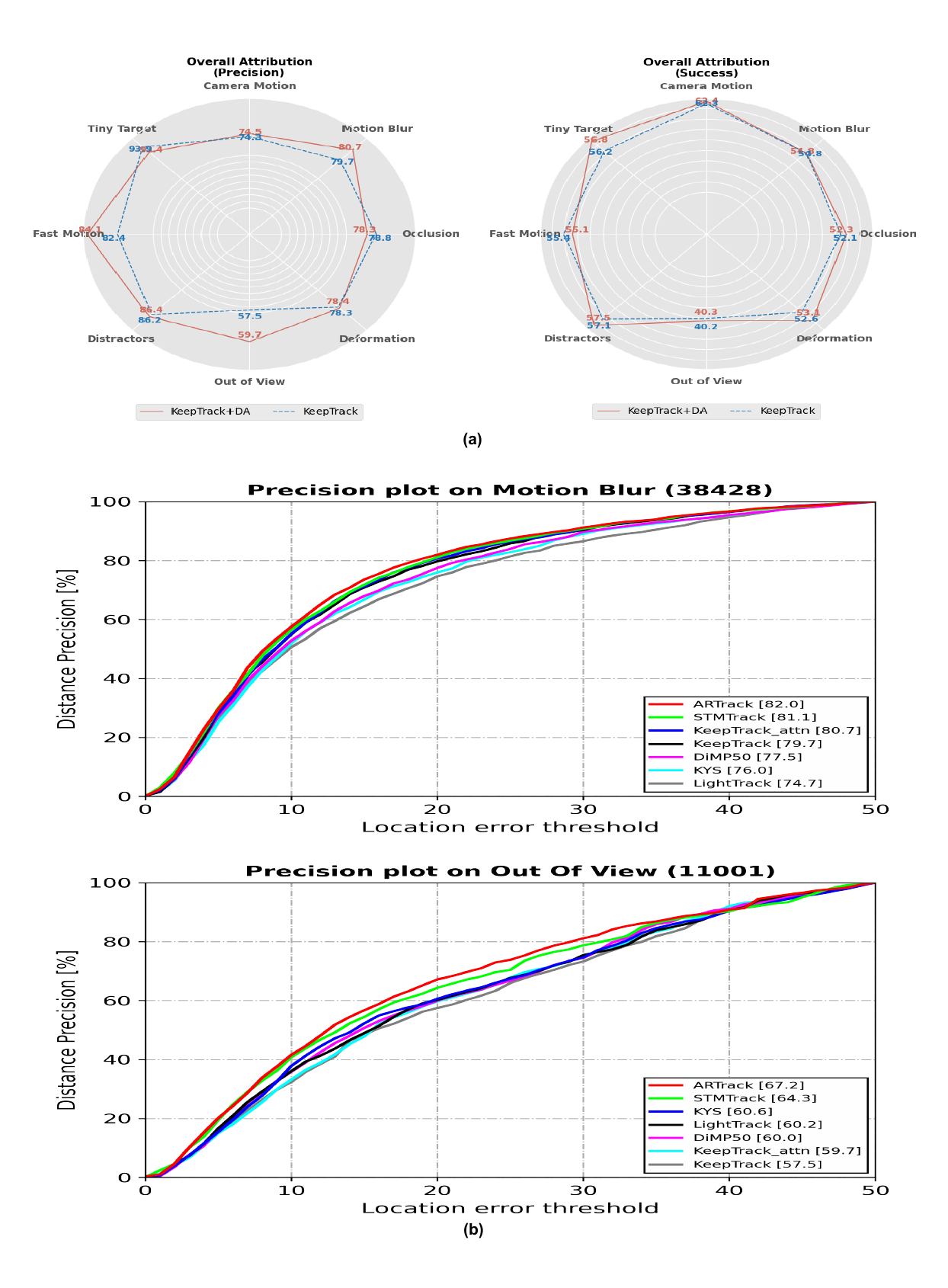}
	\caption{Evaluation of attributes on NT-VOT211. In subfigure (a), we display the attribution radar graph, comparing the original model with the model incorporating depth attention. Subfigure (b) features a detailed plot, highlighting the attributes where our module showed the most improvement, specifically Motion Blur and Out-of-View.} 
	\label{attributedanalysis}
\end{figure}

\noindent\textbf{Attributes-wise Evaluation:} 
In addition, we assessed KeepTrack+DA and other top trackers in NT-VOT211\cite{ntvot211}, with the results illustrated in Fig \ref{attributedanalysis}. This is attributes-wise evaluation. The results demonstrate that the proposed method excels in challenging scenes, such as out-of-view, motion blur, and fast motion. It tends to maintain original performance on camera motion, deformation, occlusion and distractor scenarios, while potentially exhibiting reduced performance in tiny target scenarios.

\begin{figure}[]
    \centering
    \includegraphics[width=0.80\linewidth]{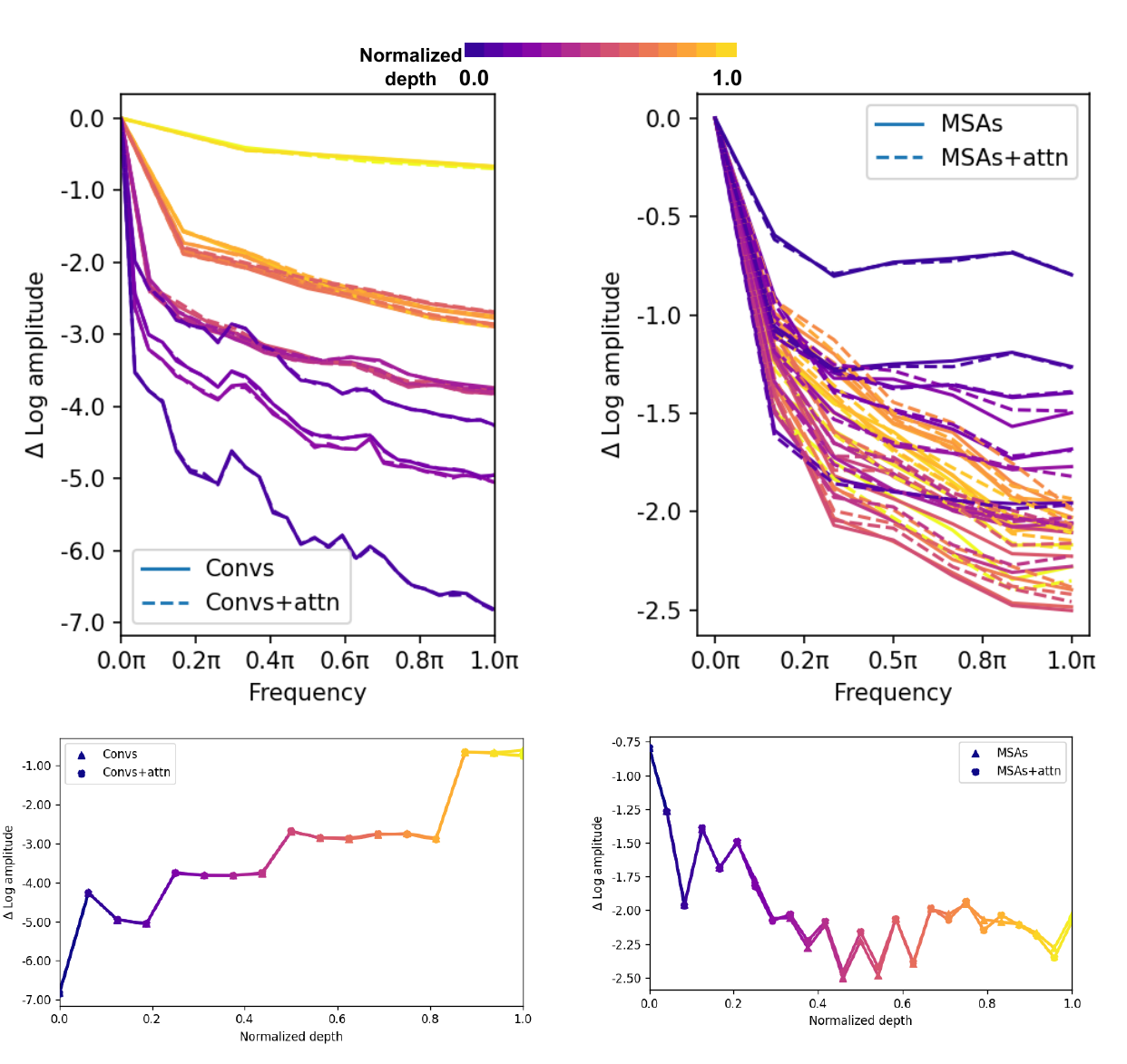}
    \caption{First row: frequency component analysis, where different depths are represented by distinct colors. A depth-indicating bar is provided at the top for reference. Second row: we present statistics on the overall amplitude corresponding to different depths.}
    \label{vit_and_conv}
\end{figure}

\noindent\textbf{Frequency Analysis:}
Inspired by the work of park et al. \cite{park2022vision}, we selected a Resnet-50\cite{he2016deep} and a ViT Tiny Patch 16/224\cite{dosovitskiy2020image} and applied the proposed method to both. Subsequently, we conducted Fourier analysis on these models, with the results depicted in Figure \ref{vit_and_conv}. It is evident from the analysis that the proposed method exhibits greater similarity to the Conv-based approach in terms of frequency analysis.
\begin{figure}[]
    \centering
    \includegraphics[width=0.80\linewidth]{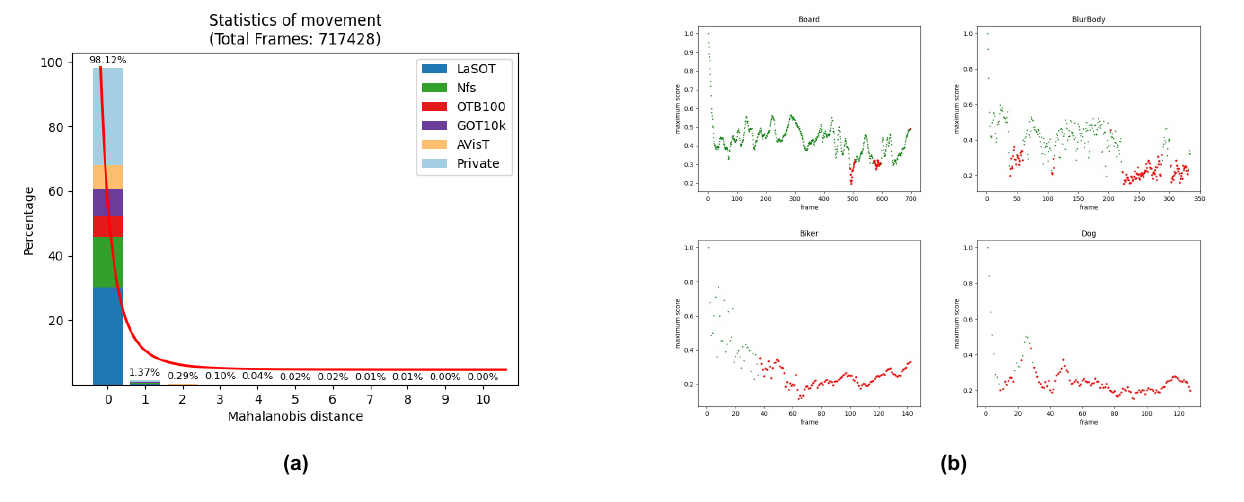}
    \caption{In subfigure a, we analyzed Manhattan Distance displacement across six benchmarks, totaling 717,428 frames. For each set of five consecutive frames, we measured the target's movement relative to its own dimensions (one unit representing length or width). Findings indicate that, in most cases, the target remained stationary, with movements not exceeding its dimensions. In subfigure b, we computed the Peak-to-Sidelobe Ratio (PSR) of the score map generated by Tracker STMTrack, showing exponential decay within a time interval. Though the algorithm may correct this in subsequent intervals, the deteriorating PSR value ultimately leads to tracker failure, marked by the red point. This highlights the need for a periodically updated prior.}
    \label{newcbfig}
\end{figure}
In the initial segments of the network, the proposed method demonstrates minimal differences, except in the vicinity around 0.2$\pi$. However, as the network progresses deeper, these distinctions become more pronounced, contributing to enhanced performance. Notably, for multi-head self-attentions (MSAs), the impact of incorporating the proposed method is particularly prominent within the interval [0.1$\pi$, 0.8$\pi$], and this effect intensifies with increasing network depth.

The observed trend in deeper networks indicates that the proposed method tends to rectify high-frequency components with larger amplitudes. In essence, the proposed method aims to attenuate the processing of the ViT-based method, preventing it from acting as a low-pass filter in the context of frequency analysis, as discussed in \cite{park2022vision}. This observation is further supported by the fact that Conv-based methods, specifically STMTrack and KeepTrack, exhibit the least improvement compared to ViT-based trackers.

\noindent\textbf{Why every five frames?} 
\label{explain2} Through statistical analysis of datasets including Nfs\cite{kiani2017need}, AVisT\cite{noman2022avist}, LaSOT\cite{fan2019lasot}, OTB100\cite{7001050}, GOT-10k\cite{huang2019got}, and NT-VOT211, we identified a distinct motion pattern for the target, as illustrated in part a of Figure \ref{newcbfig}. This pattern exhibits a long-tail distribution. This can be mathematically expressed as $\lim_{x \rightarrow 0}P(|m -x|<\epsilon) = 1$.
Where 'm' represents the movement, and it suggests that as the degree of movement within the 5 frames decreases, the likelihood of the statement being true increases. This ensures that the subsequent 4 frames can be roughly regarded as the Value Component associated with the initial frame.


\section{Conclusion and Limitations}
Our work presents a novel approach to Visual Object Tracking (VOT) by integrating monocular depth estimation in a principled way, resulting in a robust tracking system capable of handling occlusions and motion blur. We note state-of-the-art performance of our method on multiple benchmarks, including OTB100, UAV123, LaSOT, and GOT-10k. The ablation study and Fourier analysis confirm the effectiveness of our approach in overcoming occlusions and maintaining tracking continuity. However, the effectiveness of depth attention could be further enhanced through an end-to-end training approach, which is an interesting avenue for future work.

\end{document}